\documentclass[letterpaper, 10 pt, conference]{ieeeconf}  

\IEEEoverridecommandlockouts                              

\usepackage{graphics} 
\usepackage{times} 
\usepackage{amsmath} 
\usepackage{amssymb}  

\usepackage{mathtools}
\usepackage{tabularx}
\usepackage{multirow}
\usepackage{multicol}
\usepackage{graphicx}
    \graphicspath{{figure/}}
\usepackage{subcaption}
\usepackage{algorithm}
\usepackage{algpseudocode}

\newtheorem{assumption}{Assumption}
\newtheorem{theorem}{Theorem}
\newtheorem{corollary}{Corollary}

\newtheorem{definition}{Definition}

\newtheorem{problem}{Problem}

\usepackage{url}
\usepackage{xcolor}
\definecolor{mycitecolor}{RGB}{71, 191, 38}
\definecolor{mylinkcolor}{RGB}{40, 115, 201}

\makeatletter
\let\NAT@parse\undefined
\makeatother
\usepackage{hyperref}
\hypersetup{
  colorlinks=true,
  citecolor=mycitecolor,
  linkcolor=mylinkcolor,
  urlcolor=mycitecolor,
}

\title{\LARGE \bf
Integrating Predictive Motion Uncertainties with Distributionally Robust Risk-Aware Control for Safe Robot Navigation in Crowds
}

\author{Kanghyun Ryu$^{1}$ and Negar Mehr$^{1}$
\thanks{$^{1}$Kanghyun Ryu and Negar Mehr are with the Department of Mechanical Engineering, University of California, Berkeley, Berkeley, CA 94720, USA {\tt\small \{kanghyun.ryu, negar\}@berkeley.edu}}%
\thanks{This work is supported by the National Science Foundation, under grants CNS-2218759 and CCF-2211542.}
\thanks{The code is available at \href{https://github.com/labicon/DRCC-MPC.git}{https://github.com/labicon/DRCC-MPC.git}}
}

\begin{document}

\maketitle
\thispagestyle{empty}
\pagestyle{empty}

\begin{abstract}

Ensuring safe navigation in human-populated environments is crucial for autonomous mobile robots. Although recent advances in machine learning offer promising methods to predict human trajectories in crowded areas, it remains unclear how one can safely incorporate these learned models into a control loop due to the uncertain nature of human motion, which can make predictions of these models imprecise. In this work, we address this challenge and introduce a distributionally robust chance-constrained model predictive control (DRCC-MPC) which: (i) adopts a probability of collision as a pre-specified, interpretable risk metric, and (ii) offers robustness against discrepancies between actual human trajectories and their predictions. We consider the risk of collision in the form of a chance constraint, providing an interpretable measure of robot safety. To enable real-time evaluation of chance constraints, we consider conservative approximations of chance constraints in the form of distributionally robust Conditional Value at Risk constraints. The resulting formulation offers computational efficiency as well as robustness with respect to out-of-distribution human motion. With the parallelization of a sampling-based optimization technique, our method operates in real-time, demonstrating successful and safe navigation in a number of case studies with real-world pedestrian data.

\end{abstract}

\vspace{-1mm}
\section{Introduction}

\begin{figure}[t]
    \begin{subfigure}[b]{\linewidth}
        \centering
        \includegraphics[width=0.45\linewidth]{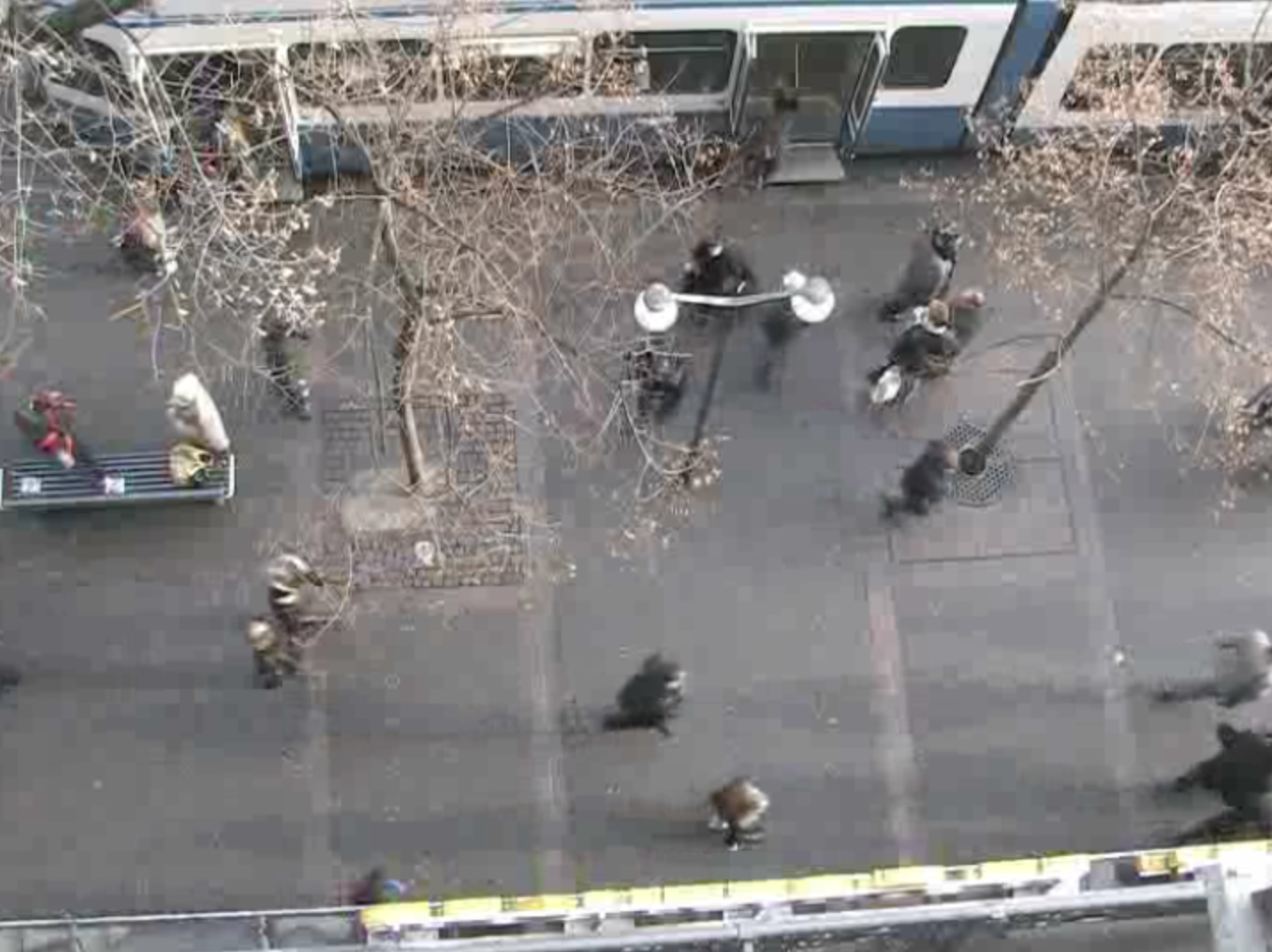}
        \caption{Real-world image of ETH/Hotel sequence.}
    \end{subfigure}

    \vspace{0.5em}
    
    \begin{subfigure}[b]{\linewidth}
        \centering
        \includegraphics[width=0.9\textwidth]{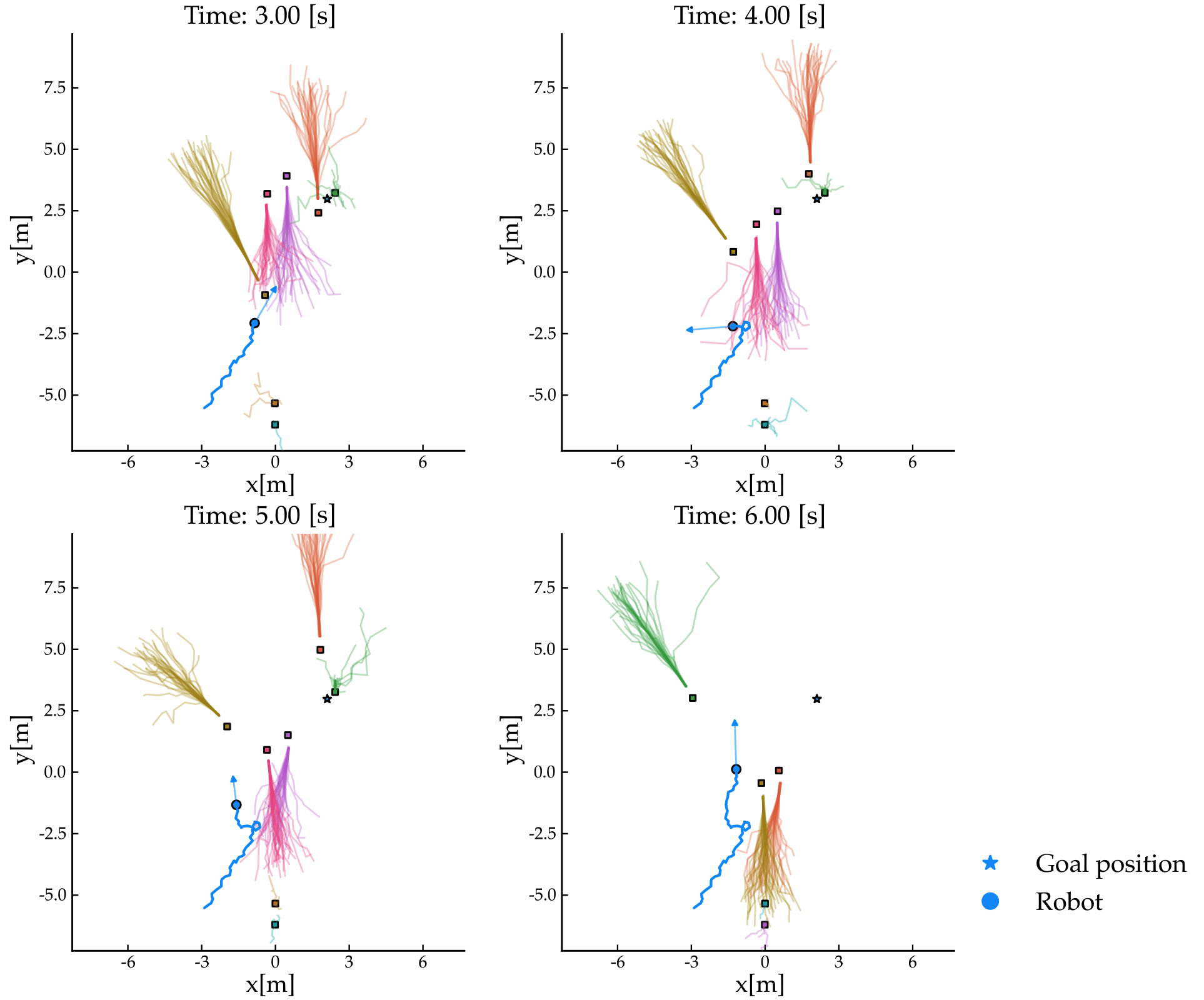}
        \caption{Example trajectory of a robot navigating among pedestrians in ETH/Hotel sequence.}
    \end{subfigure}
    \caption{Trajectory forecasting model enables a robot to estimate the distribution of future human trajectories. In response, the robot modifies its route to reduce the chance of collision.}
    \label{fig:hotel}
    \vspace{-1mm}
\end{figure}

A major challenge in incorporating autonomous robots into real-world scenarios is guaranteeing safe navigation among crowds. Humans present unique challenges due to the stochastic and potentially multi-modal, nature of their motion. However, the inherent stochasticity in human motion makes robust control approaches that account for worst-case scenarios overly cautious. Conversely, conventional stochastic control methods that focus solely on optimizing expected performance do not offer assurances of safety. To bridge this gap, in this work, we consider characterizing robot safety in the form of chance constraints which equip robots with a means to evaluate the risk associated with their actions. Notably, chance constraints also offer an intuitive interpretation of robot safety, enabling a more nuanced navigation strategy.

Evaluating chance constraints relies on understanding the distribution of human motion. Thanks to recent advancements in machine learning (ML), numerous studies explore human trajectory predictions in crowded environments~\cite{yuan2021agentformer, ettinger2021large, huang2022survey, mangalam2021goals}. Nevertheless, while these prediction modules provide reasonable models of human motion distribution, the resulting distribution of human motion is often complicated and potentially multi-modal. Hence, how to incorporate these forecasts into the robot's control loop in a safe manner remains a challenging task. In addition, their prediction may be inaccurate because of the stochastic nature of human trajectories~\cite{cheng2021limits} and the potential uncertainties stemming from sensory inputs or limited data. Relying solely on predictions, which might be imprecise, can lead to system failures, which is particularly critical in environments shared with humans. Thus, the robot's control algorithm must account for the potential inaccuracy in the human trajectory forecasts to ensure safety.

To address these challenges, we propose distributionally robust chance-constrained model predictive control (DRCC-MPC). Our proposed DRCC-MPC aims to satisfy chance constraints not only for an estimated human motion distribution but also for all human trajectory distributions within an ambiguity set. By considering distributionally robust chance constraints, we can achieve two primary enhancements: 1) it allows us to develop an approximation method for the real-time calculation of chance constraints, and 2) our controller can accommodate potential discrepancies between the actual distribution of human motion and its predicted distribution. We consider robustness against an ambiguity set to which the true distribution of human motion belongs. We construct this ambiguity set assuming that the first and second-order moments of the human motion distribution, which are estimated via Monte Carlo samples of ML-based trajectory forecasts, are known. We exploit the equivalence between chance constraints and Value-at-Risk (VaR) to create a conservative but tractable convex Conditional Value-at-Risk (CVaR) approximation of our safety constraint. Our model predictive controller employs the cross-entropy method (CEM)~\cite{botev2013cross}, a sampling-based optimization technique, to identify control sequences that meet the chance constraint. CEM optimization is parallelized using a GPU for real-time implementation. We evaluate the performance of our proposed controller in a number of navigation case studies using real-world datasets and demonstrate that our method outperforms the state-of-the-art in terms of safety in navigating human crowds. 


\section{Related Work}

\subsection{Safe Crowd Navigation}
 When a robot navigates in a crowded space with other agents, accurately modeling the behavior these agents is crucial. Earlier works employed reciprocal assumption~\cite{van2008reciprocal, van2011reciprocal} or potential fields~\cite{pradhan2011robot} for collision avoidance when other agents' velocities are given. However, they were unable to model complex interaction of agents as in crowd navigation. Social force models~\cite{yang2018social, yang2020social} try to model more complex interaction using repulsion and attraction forces based on each agent's characteristics. Still, these approaches do not account for the uncertainty in other agents' behaviors, making them less suitable for safe navigation. While game theoretic approaches have been another popular method for modeling agent interactions~\cite{williams2023distributed,wang2020game,bhatt2023efficient,Kavuncu-RSS-21}, they assume that agents have optimal policy, which may not accurately represent pedestrian movements. 

Focusing on the safety aspect of crowd navigation, Control Barrier Functions~\cite{jian2023dynamic, luo2020multi} and Hamilton-Jacobi reachability analysis~\cite{chen2016multi} can guarantee safety against the worst-case behavior of other agents. However, they tend to be overly conservative. To address this, probabilistic constraints have been proposed~\cite{andersson2016model}, assuming Gaussian distribution for the state random variable~\cite{althoff2010probabilistic, aoude2013probabilistically}. However, evaluating chance constraints for general multi-modal distributions, such as those from ML-based models, presents computationally challenges~\cite{nemirovski2007convex}.

Reinforcement learning (RL) methodologies have shown success in handling complex interactions between multiple agents by leveraging graph structures~\cite{li2020graph, liu2021decentralized} or attention mechanisms~\cite{chen2019crowd, liu2023intention, leurent2019social}. Yet, these methods often lack the interpretability and explainability required for safety assessment, which is especially crucial when deploying robots in real-world settings around humans.

Alternatively, modular approaches decouple a learning-based trajectory predictor and solve an optimal control problem for planning purposes using exhaustive search~\cite{schmerling2018multimodal} or Sequential Action Control (SAC)~\cite{nishimura2020risk}. However, they rely on soft constraints, which cannot give formal guarantees against collisions and fail to provide an interpretable notion of robot safety. Moreover, an overreliance on potentially flawed machine-learned predictions may pose significant vulnerabilities. To overcome these drawbacks, our methodology seeks interpretable safety assurance while providing robustness against distribution shifts in the trajectory forecasting module.

\subsection{Distributionally Robust Control}

Distributionally robust control (DRC) acknowledges that the true distribution of a stochastic parameter might not be accurately captured. From this observation, DRC formulates an ambiguity set, encompassing potential probability distributions of the system~\cite{rahimian2019distributionally}. The optimal control action is determined by ensuring that constraints are met for all distributions within the ambiguity set. Owing to its capacity to guarantee safety even with imperfect system modeling, DRC has been extensively employed for control in safety-critical domains~\cite{schuurmans2019safe, long2022safe}. The choice of ambiguity set is pivotal to the performance of DRC. Although several studies have defined the ambiguity set using the Wasserstein metric~\cite{hakobyan2021wasserstein} or $\phi$-divergence~\cite{sinha2020formulazero, nishimura2021rat}, determining an appropriate radius for these sets is problem-specific and not straightforward to determine in advance. As a result, moment-based ambiguity set~\cite{van2015distributionally, zymler2013distributionally} has emerged as a popular alternative to modeling moving obstacles~\cite{homchaudhuri2023distributionally}. However, their computation time complexity makes real-time operations in crowd navigation unfeasible. While Output Feedback Distributionally Robust RRT*~\cite{renganathan2020towards} presents a distributionally robust sampling-based motion planning algorithm for computational tractability, it is specifically tailored for Kalman filter-based perception modules, whereas our work can incorporate general ML-based trajectory forecasts.

\section{Preliminaries}
\label{sec:DRCC}

\subsection{Chance Constraints}
Conventional robust constraints~\cite{soloperto2019collision, bouffard2012learning} which account for worst-case uncertainties cannot handle uncertainties with unbounded support such as the stochastic pedestrian motion. In such scenarios, it is preferable to impose chance constraints that limit the probability of collision.
\begin{definition}[Chance constraint over a collision-free set] 
    For a random state $x$ where the underlying distribution is $\mathbb{P}^*$, we require 
    \begin{equation} \label{eqn:chance constraint def}
        \textnormal{Prob}^{\mathbb{P}^*}(x \in \mathcal{X}_{free}) \geq 1 - \varepsilon,
    \end{equation}
    where $\mathcal{X}_{free}$ is the collision-free space, 
 and $\varepsilon$ is the allowable probability of collision.
\end{definition}
This chance constraint provides an interpretable measure of safety requiring the system to be collision-free with at least $1 - \varepsilon$ probability. However, assessing a chance constraint, even with a known distribution, remains NP-hard~\cite{nemirovski2007convex}. We choose to approximate our chance constraint via \emph{Conditional Value-at-Risk (CVaR)}, which is well known as a tight convex approximation of chance constraints~\cite{chen2022approximations}. We reach this approximation through the notion of \emph{Value-at-Risk (VaR)} by considering a notion of a \textit{safety loss} function.
\begin{assumption} [Safety loss] \label{assumption:cost function}
    We assume that there exists a safety loss function $l:\mathbb{R}^n \rightarrow \mathbb{R}$,  which characterizes our collision-free set $\mathcal{X}_{free}$ through its zero level sets:
    \begin{equation*}
        \mathcal{X}_{free} = \{x \in \mathbb{R}^n \, |\, l(x) \leq 0\}.
    \end{equation*}
\end{assumption}
With this assumption, we can translate chance constraint~\eqref{eqn:chance constraint def} into a VaR constraint.

\begin{definition}[Value-at-Risk]
    For $x \sim \mathbb{P}^*$, and the safety loss function $l(x)$, we have:
    \begin{equation*}
\textnormal{VaR}_{\varepsilon}^{\mathbb{P}^*}(l(x)) \coloneqq \inf \{\gamma \in \mathbb{R} \, | \, \textnormal{Prob}^{\mathbb{P}^*}(l(x)>\gamma) \leq \varepsilon \}.
    \end{equation*}
\end{definition}

VaR represents the potential safety loss $l(x)$ with the allowable probability $\varepsilon$. Since chance constraint~\eqref{eqn:chance constraint def} requires $\textnormal{Prob}^{\mathbb{P}^*}(l(x) \geq 0) \leq \varepsilon$, in order for the chance constraint to be satisfied, VaR should be less or equal to zero by its definition. We can then relate the resulting VaR Constraint to Conditional Value-at-Risk (CVaR).
\begin{definition}[Conditional Value-at-Risk]
For a random state $x$ and safety loss function $l(x)$, CVaR is defined as:
\begin{equation*}
    \textnormal{CVaR}_{\varepsilon}^{\mathbb{P}^*}(l(x)) \coloneqq \inf_{\beta \in \mathbb{R}} \{\beta + \frac{1}{\varepsilon} \mathbb{E}_{\mathbb{P}^*} \{(l(x)-\beta)^+\}\},
    \end{equation*}
    where $(\cdot)^+ = \max \{\cdot, 0\}$.
\end{definition}

\begin{figure}
    \centering
    \includegraphics[width=0.6\linewidth]{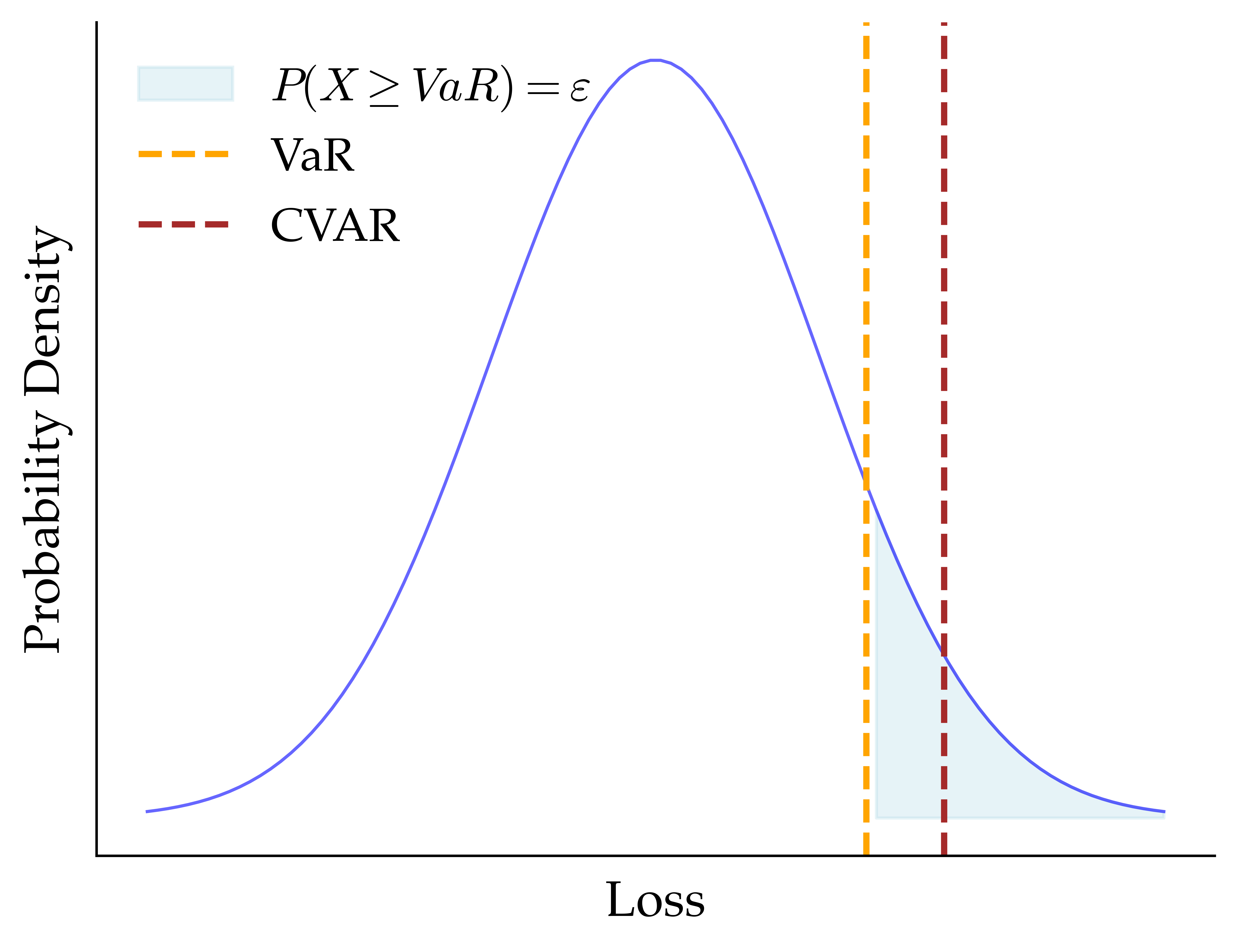}
    \caption{VaR and CVaR in a normal distribution with allowable probability $\varepsilon = 0.1$. VaR is a $1-\varepsilon$ quantile of the safety loss and CVaR is the conditional expectation of safety loss over the VaR (blue area).}
    \label{fig:CVaR}
\end{figure}

 CVaR is interpreted as the expected value over VaR, as illustrated in Figure~\ref{fig:CVaR}. Due to this trait, it naturally can be a conservative, convex approximation of VaR, effectively addressing VaR's non-convex nature~\cite{nemirovski2007convex}:
\begin{align} \label{eqn:approximation}
    \begin{split}
    \textnormal{CVaR}_{\varepsilon}^{\mathbb{P}^*}(l(x)) \leq 0 &\Rightarrow \textnormal{VaR}_{\varepsilon}^{\mathbb{P}^*} (l(x)) \leq 0 \\
    &\Leftrightarrow \textnormal{Prob}^{\mathbb{P}^*}(x \in \mathcal{X}_{free}) \geq 1-\varepsilon.
    \end{split}
\end{align}

\subsection{Distributionally Robust Constraints} \label{subsec:DRCC}

Although CVaR offers a convex approximation for chance constraints, it still requires an accurate representation of the true distribution $\mathbb{P}^*$, which can be hard to obtain for pedestrians that exhibit complex and multimodal motions. Therefore, we compute CVaR for a set of distributions $\mathcal{P}$ referred to as the \emph{ambiguity set} with the assumption that $\mathbb{P}^* \in \mathcal{P}$. Then, we consider constraint~\eqref{eqn:chance constraint def} to be distributionally robust if and only if
\begin{equation*}
    \inf_{\mathbb{P} \, \in \, \mathcal{P}} \textnormal{Prob}^{\mathbb{P}} (x \in \mathcal{X}_{free}) \geq 1-\varepsilon.
\end{equation*}
Intuitively, the above constraint requires the satisfaction of chance constraint for all possible distributions  $\mathbb{P}$ within the ambiguity set $\mathcal{P}$. Similarly, we can define distributionally robust CVaR constraint for a random state $x$ as:
\begin{equation*}
    \sup_{\mathbb{P} \, \in \, \mathcal{P}} \textnormal{CVaR}_\varepsilon^{\mathbb{P}} (l(x)) \leq 0.
\end{equation*}

Now, the connection between chance constraints, VaR, and CVaR constraints in~\eqref{eqn:approximation} can be extended to their distributionally-robust counterparts:
\begin{align} \label{eqn:dr relation}
\begin{split}
        \sup_{\mathbb{P} \, \in \, \mathcal{P}} \textnormal{CVaR}_\alpha^{\mathbb{P}} &(l(x)) \leq 0 \Rightarrow \sup_{\mathbb{P} \, \in \, \mathcal{P}} \textnormal{VaR}_\alpha^{\mathbb{P}} (l(x)) \leq 0 \\
    &\Leftrightarrow \inf_{\mathbb{P} \, \in \, \mathcal{P}} \textnormal{Prob}^{\mathbb{P}} (l(x) \leq 0) \geq 1-\alpha.
\end{split}
\end{align}

\section{Distributionally Robust Navigation in Crowded Environments}

\subsection{Notation and Problem Formulation} 

In our notation, we use subscript $r$ to refer to the robot and $i$ for the index of humans. Superscript $k$ denotes time step. Let $\mathcal{I}^k$ be the set of pedestrian indices present in $k$. Humans and a robot operate in a state space $\mathcal{X}$, while $\mathcal{X}_{robot}$ indicates the region occupied by a robot. We model the robot and humans as discrete-time dynamical systems.
\begin{subequations}
    \begin{align}
        x_i^{k+1} &= f_i(x_i^k, u_i^k), \quad \forall i \in \mathcal{I}^k, \label{eq: human dynamics} \\
        x_r^{k+1} &= f_r(x_r^k, u_r^k).    \label{eq: robot dynamics}
    \end{align}
\end{subequations}
Here, the robot does not have access to human control $u_i^k$, rendering $x_i^k$ stochastic from the perspective of the robot.

We employ a distributionally robust chance constraint to ensure robot safety. We define the system to be collision-free when all pedestrians remain outside $\mathcal{X}_{robot}$. As a result, we propose to formulate our safety constraint as:
\begin{equation} \label{eqn:safety constraint}
\begin{aligned}
     \inf_{\mathbb{P}_i^k \, \in \, \mathcal{P}_i^k} \textnormal{Prob}^{\mathbb{P}_i^k} (x_i^k \in \mathcal{X}_{i,free}^k) \geq 1-\varepsilon,\quad \forall i \in \mathcal{I}^k,
\end{aligned}
\end{equation}
where $\mathcal{X}_{i,free}^k$ represents the collision-free region for human $i$ at time step $k$.
Finally, we formulate a safe robot navigation task as a model predictive control (MPC) problem with constraint~\eqref{eqn:safety constraint}, which we call DRCC-MPC. 
\begin{problem}[DRCC-MPC] \label{problem:constrained mpc}
    \begin{subequations}
        \begin{align}
            \min_{\mathbf{x}_r, \mathbf{u}_r}& \mathcal{J} := \sum_{k=0}^{K-1} J(x_r^k, u_r^k, x_{goal}) + J_K(x_r^K, x_{goal}) \\
            s.t. \quad & x_r^{k+1} = f_r(x_r^k, u_r^k), \\
                       & x_r^k \in \mathcal{X}, u_r^k \in \mathcal{U}, \\
                       & \inf_{\mathbb{P}_i^k \, \in \, \mathcal{P}_i^k} \textnormal{Prob}^{\mathbb{P}_i^k} (x_i^k \in \mathcal{X}_{i,free}^k) \geq 1-\varepsilon, \quad \forall i \in \mathcal{I}^k, \label{eqn: problem chance constraint}
        \end{align}
    \end{subequations}
    where $\mathbf{x}_r := (x_r^0,\dots,x_r^K)$, $\mathbf{u}_r := (u_r^0,\dots,u_r^{K-1})$ are the robot state and input trajectory for horizon $K$, and $\mathcal{U}$ is the control space of the robot. In our DRCC-MPC, $J$ denotes the stage-wise cost function of the robot and $J_K$ is its terminal cost function.
\end{problem}

\subsection{Reformulation of Chance Constraints} \label{section:reformulation}

We assume that every human $i \in \mathcal{I}_k$ satisfies Assumption~\ref{assumption:cost function} for $\mathcal{X}_{i,free}^k$ with some safety loss function $l_i^k(x_i^k)$. Constraint~\eqref{eqn: problem chance constraint} can be approximated by the following distributionally robust CVaR constraint~\eqref{eqn:dr relation}, which is a sufficient condition for satisfaction of~\eqref{eqn: problem chance constraint}.
\begin{equation} \label{eqn:CVaR constraint}
    \sup_{\mathbb{P}_i^k \, \in \, \mathcal{P}_i^k} \textnormal{CVaR}_\varepsilon^{\mathbb{P}_i^k} (l_i^k(x_i^k)) \leq 0, \quad \forall i \in \mathcal{I}^k.
\end{equation}

\begin{figure}
    \centering
    \includegraphics[width=0.5\linewidth]{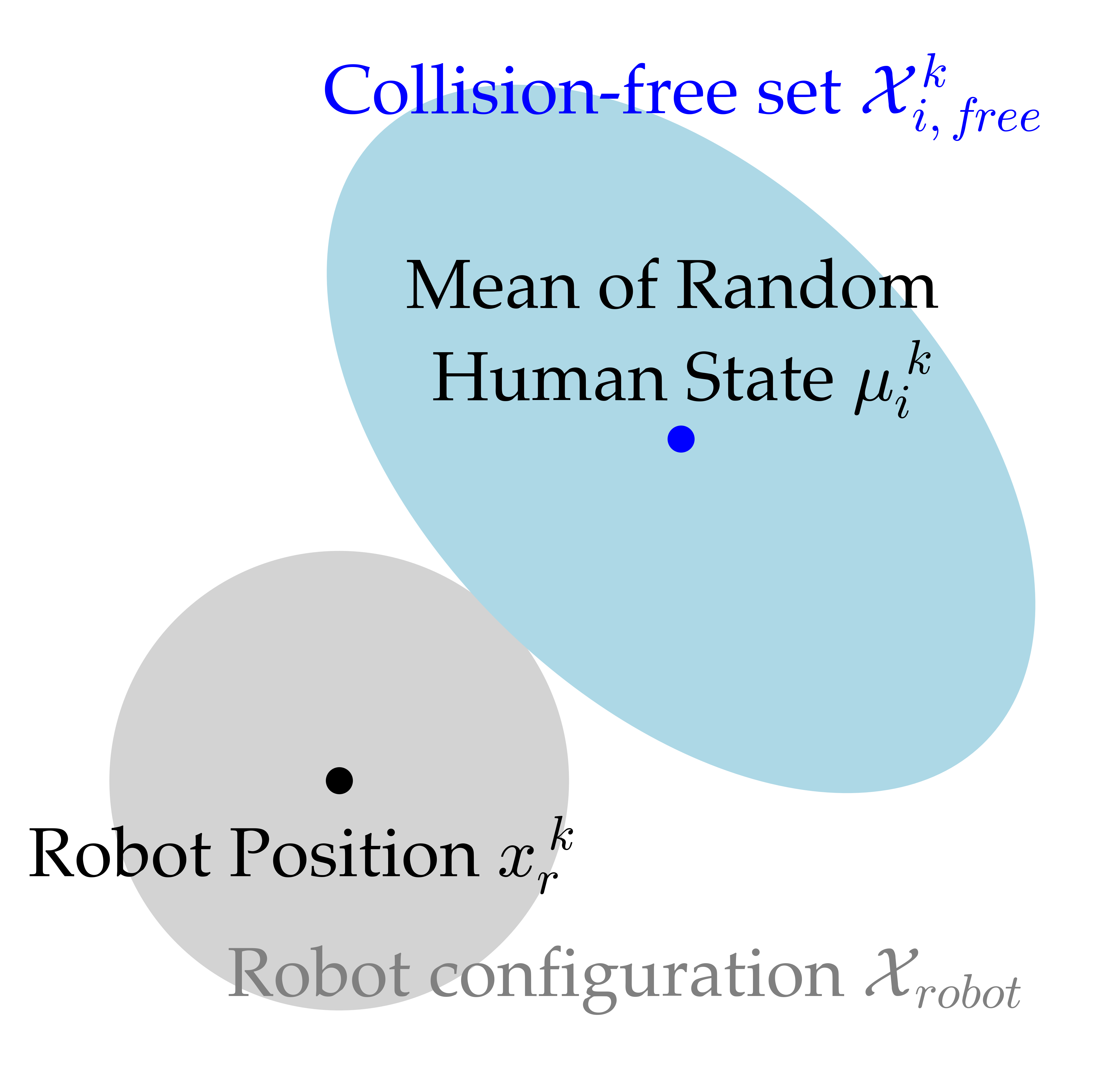}
    \caption{For human $i$, collision-free set $\mathcal{X}_{i,free}^k$ is an ellipsoid centered on $\mu_i^k$ which belongs to $\mathcal{X} \backslash \mathcal{X}_{robot}$. We constraint the random human state $x_i^k$ to belong to such a set.}
    \label{fig:safe set}
\end{figure}

To make distributionally robust CVaR constraint tractable, we require each safety loss function $l_i^k(x_i^k)$ to be convex and quadratic~\cite{van2015distributionally}, which leads to an ellipsoidal collision-free set $\mathcal{X}_{i,free}^k$. Considering that $x_i^k$ centers on its mean $\mu_i^k$, we represent $\mathcal{X}_{i,free}^k$ as an ellipsoid centered at $\mu_i^k$ outside of $\mathcal{X}_{robot}$. 
\begin{equation} \label{eqn:safe set approximation}
    \begin{aligned}
        \mathcal{X}_{i,free}^k &\coloneqq \{x \in \mathbb{R}^n | l_i^k(x_i^k) \leq 0\} \\
        &= \{x \in \mathbb{R}^n: (x-\mu_i^k)^T E_i^k (x-\mu_i^k) + e_i^k \leq 0 \},
    \end{aligned}
\end{equation}
with $E_i^k \in \mathbb{R}^{n \times n}$ where $n$ is the dimension of $x_i^k \in \mathbb{R}^n,\, E_i^k \succ 0,\, e_i^k \in \mathbb{R}$, and $\mathcal{X}_{i,free}^k \subset \mathcal{X} \backslash \mathcal{X}_{robot}$. Assuming $\mathcal{X}_{robot}$ is circle in 2D space, we can define an ellipsoid $\mathcal{X}_{i,free}^k$ whose minor axis is tangent to $\mathcal{X}_{robot}$, as illustrated in Fig~\ref{fig:safe set}.

We define the ambiguity set for the $i_{\text{th}}$ human $\mathcal{P}_i^k$ as all probability distributions $\mathbb{P}_i^k$ that have a given mean $\mu_i^k$ and covariance $\Sigma_i^k$.
\begin{equation} \label{eqn:ambiguity set}
\begin{aligned}
    \mathcal{P}_i^k \coloneqq \{\mathbb{P}_i^k | \, &\mathbb{E}_{\mathbb{P}_i^k} \{x_i^k\} = \mu_i^k, \\
                                                 &\mathbb{E}_{\mathbb{P}_i^k} \{[x_i^k - \mu_i^k]\cdot[x_i^k - \mu_i^k]^T\} = \Sigma_i^k\}.
\end{aligned}
\end{equation}   

An advantage of the moment-based ambiguity set $\mathcal{P}_i^k$ is that it requires only $\mu_i^k$ and $\Sigma_i^k$, which are easily obtained from ML-based forecasting modules with Monte Carlo sampling. Therefore, we will utilize the estimated mean $\hat{\mu}_i^k$ and covariance $\hat{\Sigma}_i^k$ of $x_i^k$ to construct an ambiguity set.

The ellipsoidal collision-free set~\eqref{eqn:safe set approximation}, combined with the moment-based ambiguity set~\eqref{eqn:ambiguity set}, makes the distributionally robust CVaR constraints tractable, which we further simplify with the following result.
\begin{theorem} \label{thm:CVaR}
    If $\mathcal{X}_{i,free}^k$ is defined as~\eqref{eqn:safe set approximation} and the random state $x_i^k$ has mean $\hat{\mu}_i^k$ and covariance $\hat{\Sigma}_i^k$, then 
    \begin{equation}
        \sup_{\mathbb{P}_i^k \, \in \, \mathcal{P}_i^k} \textnormal{CVaR}_\varepsilon^{\mathbb{P}_i^k}  (l_i^k(x_i^k)) = e_i^k + \frac{1}{\varepsilon}Tr\{\hat{\Sigma}_i^k E_i^k\},
    \end{equation}
    where $Tr\{\cdot\}$ is the trace of a matrix.
\end{theorem}
\begin{proof}
    See Proof of Corollary 1.3 in~\cite{van2015distributionally} 
\end{proof}

Theorem~\ref{thm:CVaR} provides an aid to evaluate distributionally robust CVaR constraints only using the structure of the collision-free set~\eqref{eqn:safe set approximation}, estimated mean $\hat{\mu}_i^k$, and the estimated covariance $\hat{\Sigma}_i^k$ of $x_i^k$. Therefore, we can arrive at a tractable sufficient condition for~\eqref{eqn: problem chance constraint} using~\eqref{eqn:dr relation}.
\begin{corollary} \label{corollary:general}
    If $\mathcal{X}_{i,free}^k$ is defined as~\eqref{eqn:safe set approximation} and the random human state $x_i^k$ has mean $\hat{\mu}_i^k$ and covariance $\hat{\Sigma}_i^k$, then 
    \begin{equation} \label{eqn:reformulated DRCC}
    \begin{aligned}
        e_i^k + \frac{1}{\varepsilon}Tr\{\hat{\Sigma}_i^k E_i^k\} \leq 0 &\Rightarrow \\
        \inf_{\mathbb{P}_i^k \, \in \, \mathcal{P}_i^k} \textnormal{Prob}^{\mathbb{P}_i^k} &(x_i^k \in \mathcal{X}_{i,free}^k) \geq 1-\varepsilon.
    \end{aligned}
    \end{equation}
\end{corollary}
Thus, we can reformulate Problem~\ref{problem:constrained mpc} as: 
\begin{subequations} \label{eqn:reformulated problem}
    \begin{align}
        \min_{\mathbf{x}_r, \mathbf{u}_r}& \mathcal{J} = \sum_{k=0}^{K-1} J(x_r^k, u_r^k, x_{goal}) + J_K(x_r^K, x_{goal}) \label{eqn:objective} \\
        s.t. \quad & x_r^{k+1} = f_r(x_r^k, u_r^k), \\
                   & x_r^k \in \mathcal{X}, u_r^k \in \mathcal{U}, \\
                   & e_i^k + \frac{1}{\varepsilon}Tr\{\hat{\Sigma}_i^k E_i^k\} \leq 0 , \quad \forall i \in \mathcal{I}^k, \label{eqn:problem reformulated}
    \end{align}
\end{subequations}
where $e_i^k$ and $E_i^k$ are constructed from~\eqref{eqn:safe set approximation}.

\subsection{Constrained Cross-Entropy Method}
Identifying a feasible solution for the constrained optimization problem~\eqref{eqn:reformulated problem} remains challenging, particularly due to the non-trivial process involved in retrieving $e_i^k$ and $E_i^k$ in~\eqref{eqn:problem reformulated}. To address this issue, we adopt the Cross-Entropy Method (CEM), a sampling-based optimization technique, to enable the straightforward evaluation of~\eqref{eqn:problem reformulated}. CEM recursively iterates between generating a set of robot control sequences $\{\mathbf{u}_r\}^{sampled}$ from a CEM sampling distribution and refining this distribution based on an elite set $\{\mathbf{u}_r\}^{elite}$ that achieves optimal performance. This process continues until the maximum number of iterations is reached.


As introduced in~\cite{liu2020constrained}, we should evaluate whether the sampled trajectories are feasible to solve a constrained optimization problem with CEM. Note that~\eqref{eqn:problem reformulated} can be rewritten as:
\begin{align} \label{eqn:DRCC per time step}
    \max_{i \in \mathcal{I}^k} [e_i^k + \frac{1}{\varepsilon}Tr\{\hat{\Sigma}_i^k E_i^k\}] \leq 0, \quad \forall k \in \{1,\dots,K\},
\end{align}
which indicates that if the pedestrian presenting the maximal risk satisfies the constraint, then all pedestrians will comply with the constraint. Consequently, it is computationally efficient and memory-saving to consider only the pedestrian who poses the greatest risk.

When implementing CEM, during its initial iterations, CEM might fail to sample trajectories that satisfy constraints since its sampling distribution has not yet converged. In such cases, we should select the elites set from the trajectories deemed safest. This ensures that CEM will sample safer trajectories in subsequent iterations. We use a discounted sum of~\eqref{eqn:DRCC per time step} as the \emph{Risk Score}, which evaluates the overall risk of a robot trajectory in a single scalar value.
\begin{align} \label{eqn:risk score}
    \textnormal{Risk Score}(\mathbf{x}_r) = \sum_{k=1}^K {\gamma^k \max_{i \in \mathcal{I}^k} [e_i^k + \frac{1}{\varepsilon}Tr\{\hat{\Sigma}_i^k E_i^k\}]},
    \vspace{-1mm}
\end{align}
where $\gamma$ is the discount factor. When CEM fails to sample any feasible trajectory, we will use control sequences with the lowest Risk Score as an elite set. This aids CEM to converge to less risky actions and sample safer trajectories in the next iterations. Our algorithm is summarized in Algorithm~\ref{alg:main}. In our implementation, we used Gaussian distributions for CEM sampling, where $\mu_{CEM}$ and $\sigma_{CEM}$ are the mean and standard deviation of the CEM sampling distribution. After each iteration, they are updated using the mean and standard deviation of an elite set.

\begin{algorithm} [b]
\caption{CEM optimization for DRCC-MPC} \label{alg:main}
\hspace*{\algorithmicindent} \textbf{Input} Current robot state $x_r^0$, Estimated mean $\hat{\mu}_i^{1:K}$ and covariance $\hat{\Sigma}_i^{1:K}$ of human $i \in \mathcal{I}^k$ \\
\hspace*{\algorithmicindent} \textbf{Output} Control input for robot $u_r^0$ 
\begin{algorithmic}[1]
\While{stop criteria not satisfied}
    \State $\{\mathbf{u}_r\}^{sampled} \leftarrow Sample(\mu_{CEM}, \sigma_{CEM})$
    \State $\{\mathbf{x}_r\}^{sampled} \leftarrow dynamics(x_r^0, \mathbf{u}_r^{sampled})$
    \State Evaluate constraint~\eqref{eqn:DRCC per time step}
    \State Evaluate objective function~\eqref{eqn:objective}
    \If{All trajectories not feasible}
        \State Sort with $RiskScore$~\eqref{eqn:risk score}
        \State $\{\mathbf{u}_r\}^{elite} \leftarrow SelectElite(\{\mathbf{u}_r\}^{sampled})$
    \Else 
        \State $\{\mathbf{x}_r\}^{feasible} \leftarrow SelectFeasible(\{\mathbf{x}_r\}^{sampled})$
        \State Sort with $\mathcal{J}$~\eqref{eqn:objective}
        \State $\{\mathbf{u}_r\}^{elite} \leftarrow SelectElite(\{\mathbf{u}_r\}^{feasible})$
    \EndIf
    \State $\mu_{CEM}, \sigma_{CEM} \leftarrow Update\big(\{\mathbf{u}_r\}^{elite}\big)$
\EndWhile
\State Get first control sequence in $\{\mathbf{u}_r\}^{elite}$
\State Output first control input $u^*$
\end{algorithmic}
\end{algorithm}

\section{Experiments}
\subsection{Implementation Details} 
We model our robot as a 2-D single integrator $x_r^{k+1} = x_r^k +  u_r^k \Delta t$. We employ Trajectron++~\cite{salzmann2020trajectron++} as our trajectory forecasting module for human motion. The time step of trajectory generated by Trajectron++ is $\Delta t_{Trajectron} = 0.4[s]$, while the time step for our controller and dynamics is $\Delta t = 0.1[s]$. Therefore, we use linear interpolation on the trajectory estimation to align it with our controller time step. For our evaluations, we use the ETH~\cite{pellegrini2009you} dataset, which captures various real-world human trajectories.

We consider the following quadratic objective for the robot:
\begin{equation*}
    \begin{aligned}
        J(x_r^k, u_r^k,x_{goal}) &= {\gamma^k} \big[(x_r^k - x_{goal})^T Q (x_r^k - x_{goal}) + \\
        & \hspace{4cm} (u_r^k)^T R u_r^k \big], \\
        J_K(x_r^K, x_{goal}) &= (x_r^K)^T Q_K x_r^K,
    \end{aligned}
\end{equation*}
where $Q = Q_K = 0.5I_{2 \times 2}$, $R = 0.05I_{2\times2}$, and $\gamma = 0.99$. A collision is flagged if the human-robot distance is below $r=0.4 \, [m]$. We consider the planning horizon to be $K=40$. The robot operates within a maximum input bound $u_{max} = 2.0 \, [m/s]$. $M=30$ trajectories are drawn from the Trajectron++ module to construct ambiguity set. During the $5$ iterations of CEM optimization, we sample $400$ control sequences and select $40$ elite set. CEM optimization used GPU parallel computing to achieve real-time performance in a desktop computer with an AMD Ryzen 5 5600 CPU and an NVIDIA GeForce RTX 3060 GPU.

We use the Buffered Input Cell (BIC) method~\cite{wang2018safe} as a reference for reciprocal collision avoidance and the CrowdNav~\cite{chen2019crowd} model as RL baseline. Among the modular control strategies, we considered exhaustive search with motion primitives~\cite{schmerling2018multimodal} and Risk-Sensitive Sequential Action Control (RSSAC)~\cite{nishimura2020risk} since they both account for the stochasticity of human behavior. They incorporate a collision cost, measured by the distance between humans and the robot within their optimization objective. While exhaustive search and RSSAC with risk-sensitive parameter $\sigma = 0$ identifies the optimal control by assessing average cost, RSSAC with $\sigma > 0$ optimizes the entropic risk measure~\cite{majumdar2020should}. The codes and hyperparameters for these baselines are from RSSAC's code repository. Lastly, we consider different collision probabilities $\varepsilon = 0.05,\, 0.1,\, 0.15,$ to observe the effect of the collision probability $\varepsilon$ on performance of our method.

\subsection{Real-World Pedestrian Dataset}

Our evaluation using the ETH dataset considers two unique sequences: the "ETH/Hotel" and "ETH/ETH" sequences. Over a span of 10 seconds, the Hotel sequence has a total of 8 pedestrians, whereas the ETH sequence includes 16 pedestrians. Detailed results can be found in Table~\ref{table:main result}.

\begin{table*}[t]
    \centering
    \begin{subtable}{\linewidth}
        \centering
        \caption{Hotel sequence}
        \label{subtable:hotel}
        \begin{tabular}{|c|c|c|c|c|c|}
            \hline
            \textbf{Method} & \textbf{Collision} & \textbf{Minimum Distance}  & \textbf{Success} & \textbf{Positional Cost} & \textbf{Computation Time(ms)} \\
            \hline
            BIC & 2.7986 $\pm$ 3.6774 & 0.4253 $\pm$ 0.3208 & 48.46 & 204.19 $\pm$ 153.25 & 2.80 $\pm$ 5.17\\
            \hline
            CrowdNav & 0.3368 $\pm$ 1.2555 & 1.8846 $\pm$ 1.7901 & 92.59 & 617.61 $\pm$ 629.79  & 59.10 $\pm$ 6.02\\
            \hline
            Exhaustive& 0.0519 $\pm$ 0.3504 & 1.2522 $\pm$ 0.3287 & 96.59 & 240.55 $\pm$ 168.66 & 273.57 $\pm$ 101.78\\
            \hline
            RSSAC($\sigma = 0$) & 0.0109 $\pm$ 0.1330 & 0.9473 $\pm$ 0.1547 & 99.32 & 196.04 $\pm$ 120.09 & 24.08 $\pm$ 22.23\\ 
            RSSAC($\sigma = 1$) & 0.0403 $\pm$ 0.2687 & 0.9645 $\pm$ 0.2351 & 97.27 & 242.33 $\pm$ 137.57 & 23.40 $\pm$ 21.62\\
            \hline
            DRCC-MPC($\varepsilon = 0.05$) & \textbf{0.0000 $\pm$ 0.0000} & 0.7166 $\pm$ 0.1874 & \textbf{100} & 239.68 $\pm$ 142.21 & 39.58 $\pm$ 22.03\\ 
            DRCC-MPC($\varepsilon = 0.1$) & \textbf{0.0102 $\pm$ 0.1008} & 0.5706 $\pm$ 0.0869 & 98.97 & 198.83 $\pm$ 122.80 & 39.66 $\pm$ 22.04 \\ 
            DRCC-MPC($\varepsilon = 0.15$) & \textbf{0.0068 $\pm$ 0.0825} & 0.5312 $\pm$ 0.0809 & \textbf{99.32} & \textbf{184.89 $\pm$ 116.67} & 39.72 $\pm$ 23.16\\ 
            \hline
        \end{tabular}
    \end{subtable}
    

    \begin{subtable}{\linewidth}
        \centering
        \caption{ETH sequence}
        \label{subtable:eth}
        \begin{tabular}{|c|c|c|c|c|c|}
            \hline
            \textbf{Method} & \textbf{Collision} & \textbf{Minimum Distance}  & \textbf{Success} & \textbf{Positional Cost} & \textbf{Computation Time(ms)} \\
            \hline
            BIC & 2.5717 $\pm$ 2.2883 & 0.4030 $\pm$ 0.3512 & 31.31 & 255.96 $\pm$ 186.43 & 2.38 $\pm$ 1.41 \\
            \hline
            CrowdNav & 0.4878 $\pm$ 1.7498 & 1.5893 $\pm$ 1.4401 & 91.25 & 599.36 $\pm$ 491.81 & 59.51 $\pm$ 5.60\\
            \hline
            Exhaustive& 0.0815 $\pm$ 0.3328 & 1.0024 $\pm$ 0.3988 & 91.92 & 232.72 $\pm$ 169.18 & 455.46 $\pm$ 193.66\\
            \hline
            RSSAC($\sigma = 0$) & 0.0154 $\pm$ 0.1886 & 1.0893 $\pm$ 0.5462 & 99.33 & \textbf{169.62 $\pm$ 144.92} & 36.99 $\pm$ 44.48 \\ 
            RSSAC($\sigma = 1$) & 0.0208 $\pm$ 0.2604 & 1.0589 $\pm$ 0.4661 & 99.33 & 195.34 $\pm$ 169.11 & 35.76 $\pm$ 49.52\\
            \hline
            DRCC-MPC($\varepsilon = 0.05$) & \textbf{0.0034 $\pm$ 0.0580} & 0.8690 $\pm$ 0.3310 & \textbf{99.66} & 285.41 $\pm$ 193.76 & 51.33 $\pm$ 31.53\\ 
            DRCC-MPC($\varepsilon = 0.1$) & \textbf{0.0000 $\pm$ 0.0000} & 0.6926 $\pm$ 0.2723 & \textbf{100} & 223.97 $\pm$ 168.24 & 51.31 $\pm$ 31.43\\ 
            DRCC-MPC($\varepsilon = 0.15$) & \textbf{0.0067$ \pm$ 0.0819} & 0.6878 $\pm$ 0.3770 & \textbf{99.33} & 201.52 $\pm$ 156.01 & 50.59 $\pm$ 31.71 \\ 
            \hline
        \end{tabular}
    \end{subtable}
    \caption{Simulation result from 300 experiments with different initial and goal states. Average and standard deviation is reported while excluding experiments with initial state collisions. DRCC-MPC achieved safer results compared to other baselines.``Collision": Represents the collision rate normalized over a 10-second duration. We report the normalized value because of the different control frequencies across methods. "Minimum Distance": Minimum distances between the human and robot throughout the navigation. ``Success": Percentage of experiments that were concluded without a collision. ``Positional Cost": Sum of the squared distances to the goal throughout the trajectory $\sum_{k = 0}^{end} ||x_r^k - x_{goal}||^2 \Delta t$. ``Computation Time": Wall time of each method for control and prediction update.}
    \vspace{-2em}
    \label{table:main result}
\end{table*}

\begin{figure}[t]
\centering
\includegraphics[width=0.35\textwidth]{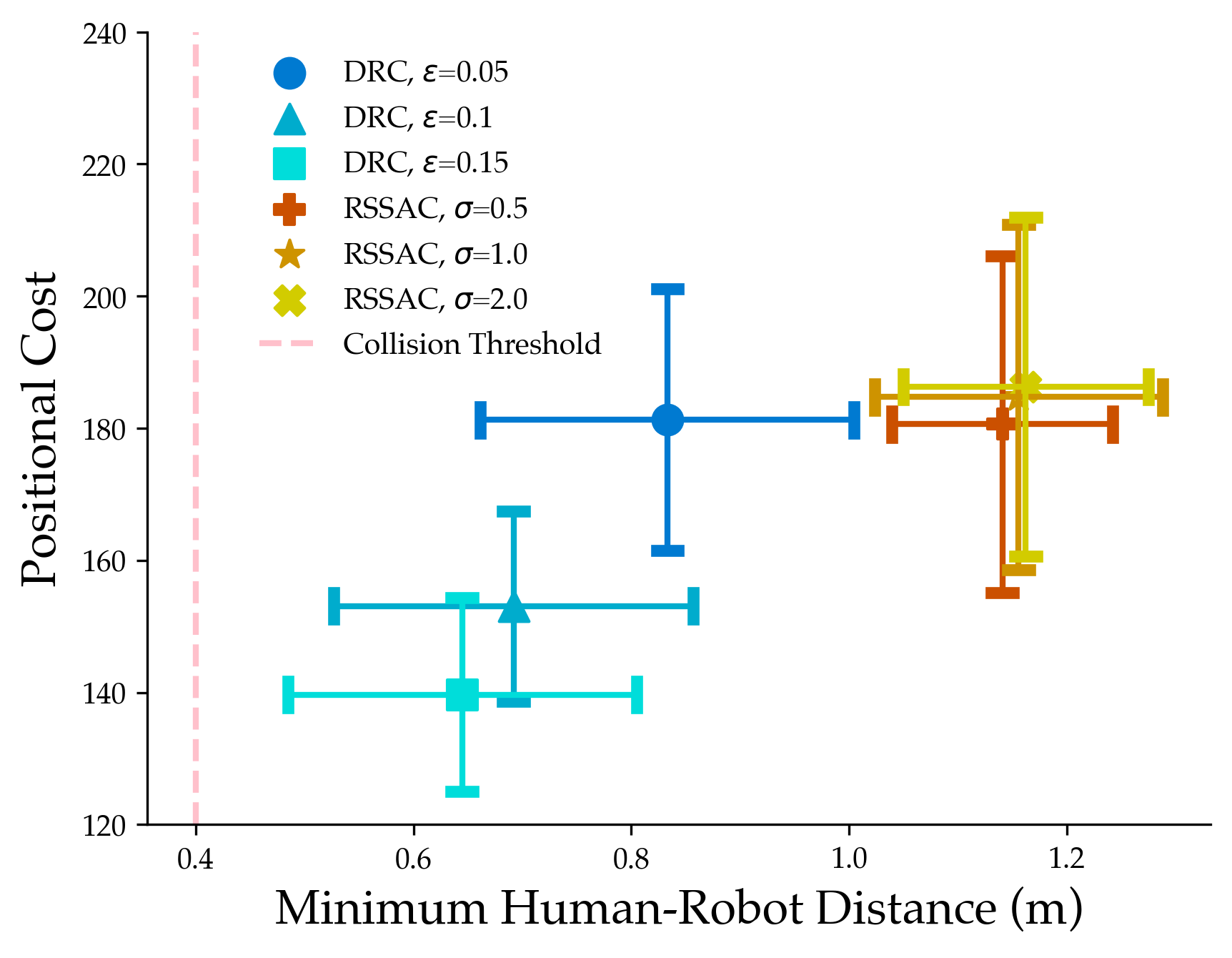}
\caption{Sensitivity analysis in Hotel sequence with 100 experiments. As the DRCC-MPC becomes more conservative by reducing $\varepsilon$, the robot tends to keep a larger distance from the humans and decides to take a longer path. On the contrary, RSSAC showed similar behavior for different choices of its risk-sensitivity parameter $\sigma$, showing an inability to display a straightforward interpretation of risk and safety.}
\label{fig:hotel_sensitivity}
\end{figure}

In both sequences, DRCC-MPC outperformed the baseline methods in reducing collisions and improving success rate. We note that the poor performance of BIC is attributed to the violation of its reciprocal assumption, primarily because we relied on a fixed real-world pedestrian dataset. This implies that BIC highly relies on a reciprocal assumption, which might not be true in a crowd-dense environment. CrowdNav often failed to reach the goal, resulting in a significant performance cost. A potential reason for this could be its initialization through a suboptimal policy~\cite{liu2021decentralized}, making it challenging to adapt to more complex scenarios. Exhaustive search with $9^4 = 6561$ motion primitives failed to run in real-time. Notably, DRCC-MPC ensured fewer collisions and maintained a closer minimum distance than RSSAC, the best-performing baseline. This is because RSSAC's collision cost includes an added margin to ensure safety. In addition, the risk sensitivity parameter $\sigma$ in RSSAC did not show a significant difference in navigation safety and it only resulted in higher positional costs compared to the risk-neutral setting. This is because $\sigma$ is more oriented towards determining yielding behaviors to pedestrians~\cite{nishimura2020risk}. This interpretation can be non-intuitive, especially when trying to map it directly to real-world safety. In contrast, our collision probability $\varepsilon$ in DRCC-MPC is more straightforward and can be directly related to navigation scenarios, offering a clearer and potentially more reliable safety parameter. This confirms our hypothesis that chance constraints provide a more intuitive and interpretable measure of risk.
\vspace{-0.1mm}
\subsection{Sensitivity to Chance Constraint Probability}
Now, we investigate the impact of collision probability $\varepsilon$ on performance. The results from Table~\ref{table:main result} clearly illustrate the safety-efficiency trade-off determined by the choice of $\varepsilon$. For instance, in the Hotel sequence, selecting $\varepsilon = 0.05$ over $\varepsilon = 0.15$ led to a 34\% increase in the minimum distance. In the ETH sequence, this increment was 26\%. However, opting for the smaller $\varepsilon$ value also resulted in substantial increase in positional cost: 29\% in the Hotel sequence and 41\% in the ETH sequence. 

For a more in-depth examination of the effect of $\varepsilon$, we carried out a controlled experiment, holding the robot's starting and goal positions constant. Fig~\ref{fig:hotel_sensitivity} illustrate the mean and standard deviation of minimum human-robot distance and positional cost, for various $\varepsilon$. We observed a direct relationship between $\varepsilon$ and minimum human-robot distance; smaller $\varepsilon$ increased the minimum human-robot distance. Conversely, a larger $\varepsilon$ led the robot to navigate through riskier paths, thereby reducing the positional cost. As a result, the direct relationship between $\varepsilon$, trajectory uncertainty, and safe distance becomes readily comprehensible. This relationship provides system designers with the significant advantage of being able to adjust the safety-performance trade-off easily. This contrasts to the risk sensitivity parameter $\sigma$ in RSSAC, which did not influence the outcome of the safety evaluation.
\vspace{-1.3em}
\section{Conclusions and Future Works}

In this paper, we addressed the problem of safe robot navigation in the presence of moving pedestrians. We incorporated a distributionally robust chance constraint as a risk metric in our MPC problem. We modeled a moment-based ambiguity set from Monte Carlo samples of human trajectory predictions. Then, distributionally robust chance constraint was reformulated using an ellipsoidal collision-free set. Finally, a control sequence that satisfied the proposed constraint was found with CEM, which was implemented with GPU for real-time computation. Our control algorithm successfully demonstrated the ability to navigate safely using a human motion forecaster which was trained on a real-world dataset. Looking forward, we intend to extend our work to environments where humans change their behavior in response to the robot's action.

\addtolength{\textheight}{-0cm}   

\bibliographystyle{ieeetr}
\bibliography{literature}

\end{document}